\documentclass[a4paper,conference]{IEEEtran}
\usepackage{graphicx}

\begin{document}
\title{ComDensE : Combined Dense Embedding of Relation-aware and Common Features for Knowledge Graph Completion}

\author{\IEEEauthorblockN{Minsang Kim\textsuperscript{\rm 1,\rm 2},  Seungjun Baek\textsuperscript{\rm 2}}
Email: kmswin7@gmail.com, sjbaek@korea.ac.kr\\
\textsuperscript{\rm 1}Kakao Enterprise, \textsuperscript{\rm 2}Korea University
}

\maketitle

\begin{abstract}
Real-world knowledge graphs (KG) are mostly incomplete. The problem of recovering missing relations, called KG completion, has recently become an active research area. Knowledge graph (KG) embedding, a low-dimensional representation of entities and relations, is the crucial technique for KG completion. Convolutional neural networks in models such as ConvE, SACN, InteractE, and RGCN achieve recent successes. This paper takes a different architectural view and proposes ComDensE which combines relation-aware and common features using dense neural networks. In the relation-aware feature extraction, we attempt to create relational inductive bias by applying an encoding function specific to each relation. In the common feature extraction, we apply the common encoding function to all input embeddings. These encoding functions are implemented using dense layers in ComDensE. ComDensE achieves the state-of-the-art performance in the link prediction in terms of MRR, HIT@1 on FB15k-237 and HIT@1 on WN18RR compared to the previous baseline approaches. We conduct an extensive ablation study to examine the effects of the relation-aware layer and the common layer of the ComDensE. Experimental results illustrate that the combined dense architecture as implemented in ComDensE\footnote{ https://github.com/kmswin1/ComDensE} achieves the best performance.
\end{abstract}
\IEEEpeerreviewmaketitle

\section{Introduction}

\noindent The knowledge graph (KG) is an essential tool to represent real-world facts and related knowledge. The KG is defined as a directed heterogeneous graph; KGs contain nodes and edges with types and directions, where nodes represent entities and edges represent their relationships. To represent a relation between entities, triples $(s, r, o)$ of KGs are defined where $s$, $r$ and $o$ represent ``subject-entity'', ``relation'' and ``object-entity'' respectively.

Recently, KGs have been widely adopted in applications for language models like LUKE~\cite{yamada2020luke}, KALM~\cite{rosset2020knowledge}, KELM~\cite{agarwal2020knowledge}, KGBART~\cite{liu2020kg}; question answering models like Query2Box~\cite{ren2020query2box}, CQD~\cite{arakelyan2020complex}, LEGO~\cite{ren2021lego}; information extraction models which extract knowledge graph from text such as Distant Supervision for Relation Extraction~\cite{mintz2009distant}, Knowledge Vault~\cite{dong2014knowledge}, End-to-End Neural Entity Linker~\cite{shang2019end}, Autoregressive Entity Retrieval~\cite{de2020autoregressive}.

\begin{figure}[t]
    \centering
    \includegraphics[width=3in]{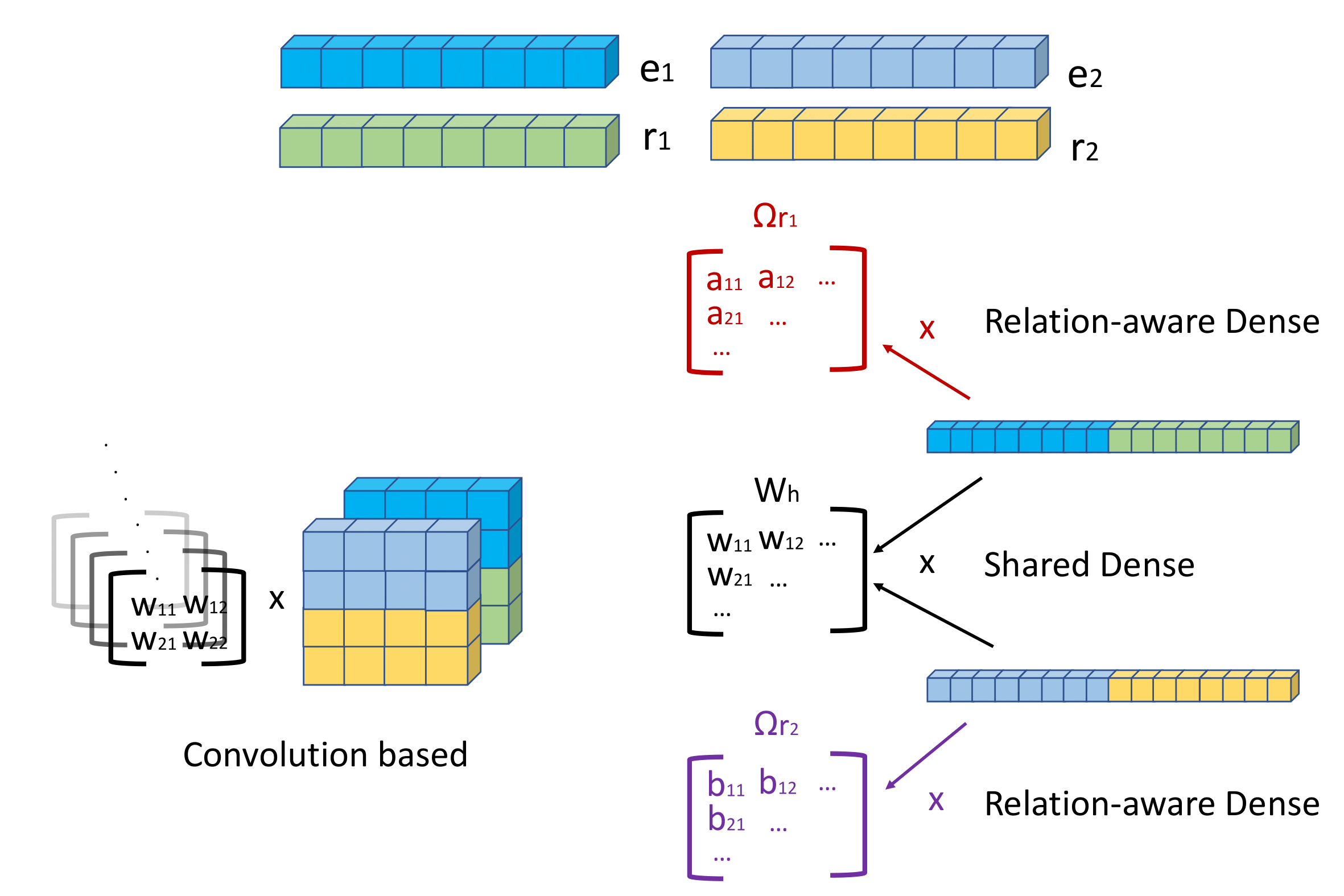}
    \caption{ComDensE combines relation-aware dense layers specific to the relation types contained in triples, and common dense layers shared among all the triples. Convolution (left) and ComDensE (right). $e_i$ denote an entity embedding vector and $r_i$ denote a relation embedding vector.}
\label{fig1}
\end{figure}

However, popular KGs such as WordNet~\cite{miller1995wordnet}, FreeBase~\cite{bollacker2008freebase}, Dbpedia~\cite{lehmann2015dbpedia}, and YAGO~\cite{rebele2016yago} are mostly \emph{incomplete}. Thus, the task of completing missing relations among entities, called the knowledge graph (KG) completion, has become a active research area. A related technique is the KG embedding which finds the representation of entities and relations in low-dimensional continuous vector spaces. One of the foundational approaches for KG embeddings is based on \textit{translational distance}~\cite{shang2019end}, which performs embeddings on Euclidean or non-Euclidean space, and uses translational distance between embeddings based scoring functions for link prediction. Those works include TransE~\cite{bordes2013translating}, TransH~\cite{wang2014knowledge}, TransD~\cite{ji2015knowledge} and RotatE~\cite{sun2019rotate}. Another type of KG embedding is the \textit{non-neural semantic matching model} such as DistMult~\cite{yang2014embedding}, ComplEx~\cite{trouillon2016complex}, HolE~\cite{nickel2016holographic}, which propose embeddings on Euclidean or non-Euclidean spaces, and use similarity based scoring functions capturing more complex semantic information. 

Recently, \textit{neural network based models} have been introduced. The models based on convolutional neural networks (CNN) use 2D reshaped embeddings, e.g., ConvE~\cite{dettmers2018convolutional},  ConvTransE~\cite{shang2019end}, SACN~\cite{shang2019end}, ReInceptionE~\cite{xie2020reinceptione} and InteractE~\cite{vashishth2020interacte}. RGCN~\cite{schlichtkrull2018modeling} exploits the connectivity structures of KGs using graph convolutional network (GCN). The CNN based approaches have yielded outstanding results on link prediction.

In this paper, we consider KG embedding using neural networks, but begin by questioning the use of convolution operations over embeddings. Somewhat surprisingly, we find that a proper design using \emph{dense} or fully connected networks combining relation-aware and common features suffices to achieve outstanding link prediction performance. The rationale behind our approach is as follows.

\textbf{Combined relation-aware and common features.} We first propose to combine the \emph{relation-aware encoding functions} and the \emph{common encoding functions}. The motivation of combining two layers is that since KGs contain diverse relation types, and the number of relations is typically less than the number of entities in KGs (e.g., 11 vs 40K in WN18RR, 237 vs 14K in FB15k-237), the relation-aware encoder, which applies to specific relations, better discriminates each relational feature. On the other side, the common encoder, which is composed of a wide dense matrix, captures various features from all triples like multiple filters in CNN. Experimental results demonstrate that combining two layers illustrates a synergistic effect.

\textbf{Dense vs Convolution.} Previous approaches using CNN applied convolution operations to concatenated (and reshaped) embeddings of entities and relations. The key properties of convolution operations include locality and translational invariance. However, unlike processing image features, the benefits of these properties applied to KG embeddings are unclear. Instead, a filtering operation applied globally to the concatenated embeddings can fully combine the feature components from entity and relation embeddings. Dense layers enable such a global filtering, because the kernel of a dense layer covers the entire feature map of embeddings. Moreover, KG embeddings are not very high-dimensional, unlike high-resolution image features. Thus, dense networks for KG embeddings incurs manageable computational overhead such as parameter count, which we show by experiments.

\textbf{Width vs Depth.} We posit that the success of convolutional methods is attributable to the use of multiple filters in convolutional layers. We make a similar observation: the \emph{width} of dense layer which corresponds to the number of stacked fully connected filters, is a key design parameter. Experiments show that the performance can be significantly improved by optimizing the width. By contrast, increasing \emph{depth} of the neural network can be harmful.

We combine these ideas and propose \emph{ComDensE}, a simple dense neural architecture for KG embedding which combines both relation-aware and common features among embeddings, as in Fig. \ref{fig1}. Our contributions are summarized as follows.

\begin{itemize}
	\item We propose ComDensE, a simple KG embedding architecture that utilizes dense layers for combining the extractions of relation-aware and common features from embeddings. ComDensE exploits inductive biases specific to relation types and provides full-sized filtering of concatenated embeddings of entities and relations. In addition, ComDensE is easy to tune, i.e., the main hyperparameter of the model is the width of the common dense layer.

    \item We evaluate link prediction performances with WN18RR and FB15k-237 datasets. Experimental results demonstrate that CombinE outperforms previous neural network methods, and achieves state-of-the-art MRR, HIT@1 on FB15k-237 and HIT@1 on WN18RR.
    
    \item We extensively conduct ablation study to support our claims. We experiment the model with common feature extraction only, adjusting both width and depth. The study confirms that ComDensE, which combines the both models, achieves the best performance.
\end{itemize}

\begin{figure}[t]
    \centering
    \includegraphics[width=3.5in]{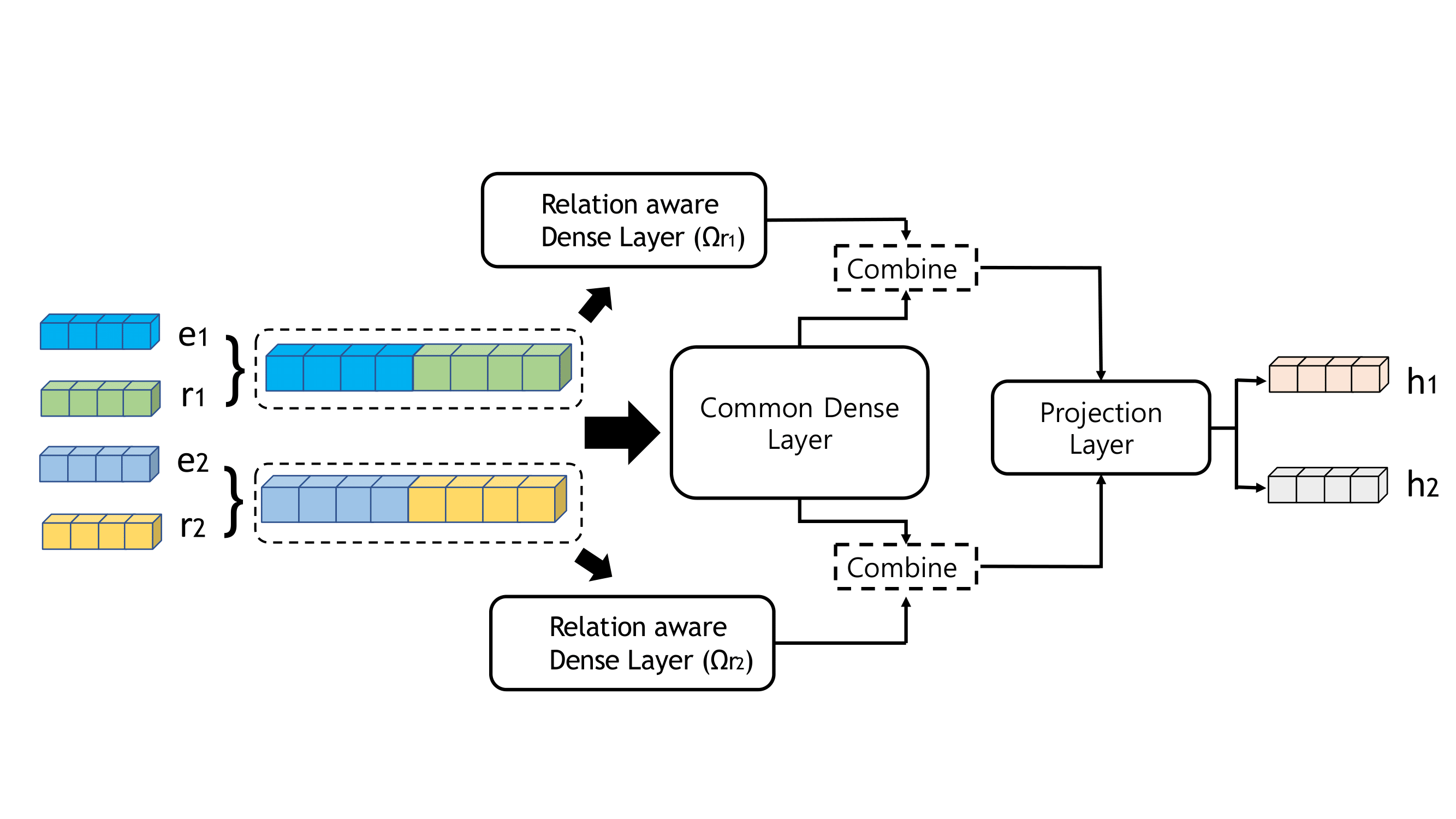}
    \caption{ComDensE model architecture.}
    \label{fig2}
\end{figure}

\section{Related Work}
\textbf{Translational Distance Models:} The knowledge graph embedding for link prediction became an active research area. The KG embedding was initiated by TransE~\cite{bordes2013translating}, which is based on the translation, i.e., the vector addition, of entity and relation embeddings in a low-dimensional vector space. TransH~\cite{wang2014knowledge} is another distance-based model which uses translation on relation-specific hyperplanes through the orthogonal projection. RotatE~\cite{sun2019rotate} considers embeddings on the complex space, and proposes to use Euler's formula to represent relations as rotation on a unit circle.\\
\indent \textbf{Non-neural Semantic Matching Models:} DistMult~\cite{yang2014embedding} is an embedding model in the Euclidean space, and computes a prediction score using bilinear functions. ComplEx~\cite{trouillon2016complex} is a semantic matching model where all entities and relation embeddings are on the complex space.\\
\indent \textbf{Neural Network based Models:} Neural Tensor Network (NTN)~\cite{socher2013reasoning} proposes to use a relation-specific transformation. Recently, embedding models based on Graph Convolutional Network (GCN)~\cite{kipf2016semi} have been proposed so as to exploit the connectivity structures of KGs. RGCN~\cite{schlichtkrull2018modeling} is a GCN based method leveraging the adjacency matrix of knowledge graph and relational weight matrices. RGCN introduces basis and block-diagonal decomposition methods in order to prevent an over-fitting. CNN have been widely used for the KG embedding. ConvE~\cite{dettmers2018convolutional} applies convolutional filters to 2D-reshaped subject and relation embeddings. ConvE computes similarity scores between obtained hidden vectors for link prediction. SACN~\cite{shang2019end} is an end-to-end architecture using weighted GCN as the encoder, and Conv-TransE, or ConvE with translational characteristics, as the decoder. InteractE~\cite{vashishth2020interacte} takes a similar approach to ConvE, but uses feature permutation and a circular convolution operation so as to enhance interactions among the components of embedding vectors.

\begin{table}
\centering
\caption{Scoring functions $\psi(e_s, e_r, e_o)$ of KG embedding methods. $e_s, e_r, e_o \in \Re^d$ for TransE, DistMult, ConvE, InteractE, and ComDensE, $e_s, e_r, e_o \in \mathrm{C}^d$ for ComplEx and RotatE. $\circ$ denote Hadamard-product, * denote convolution, $\bullet$ denote circular convolution, $\Psi(\mathcal{P}_k)$ denote feature permutation, $f$ and $g$ denote activation functions, $[\bar{e_s};\bar{e_r}]_{2d}$ represent 2d reshaping, $[e_s;e_r]_{1d}$ represent vector concatenation, $\Omega$ denote shared dense layer and $\Omega_r$ indicate relation-aware operation.}
\label{table1}
\begin{tabular}{c|c}
    \textbf{Model} & \textbf{Scoring function $\psi(e_s, e_r, e_o)$}\\
    \hline
    TransE & $||e_s + e_r - e_o||_p$\\
    RotatE & $-||e_s \circ e_r - e_o||^2$\\ 
    DistMult & $<e_s, e_r, e_o>$\\
    ComplEx & $Re(<e_s, e_r, e_o>)$\\
    ConvE & $f(vec(f([\bar{e_s};\bar{e_r}]_{2d} * w))W)e_o$\\
    InteractE & $g(vec(f(\Psi(\mathcal{P}_k) \bullet w))W)e_o$\\
    \hline
    ComDensE & $f([f(\Omega_r([e_s;e_r]_{1d})); f(\Omega([e_s;e_r]_{1d}))]_{1d}^TW)e_o$
\end{tabular}
\end{table}

\begin{table*}[t]
\centering
\caption{Statistics of benchmark datasets. PageRank of datasets is cited from~\cite{dettmers2018convolutional}.}
\label{table2}
\begin{tabular}{c|c|c|c|c|c|c}
 \textbf{Dataset} &
 \textbf{\#Entities} &
 \textbf{\#Relations} &
 \textbf{\#Training triples} &
 \textbf{\#Validation triples} &
 \textbf{\#Test triples} &
 \textbf{PageRank} \\
 \hline
 FB15k-237 & 14,541 & 237 & 272,115 & 17,535 & 20,466 & 0.733 \\
 WN18RR & 40,943 & 11 & 86,835 & 3,034 & 3,134 & 0.104
 \\
\end{tabular}
\end{table*}

\section{Method}
This section describes the details of ComDensE architecture which is depicted in Fig.~\ref{fig2}. Let $e_s, e_o \in \Re^{d_e}$ denote the embedding vectors of subject and object entities, and $e_r \in \Re^{d_r}$ denote the embedding vector of relations. In order to compute the scoring function of ComDensE, we begin by concatenating the subject and relation vectors as follows:
\begin{equation}
[e_s;e_r]_{1d} \in \Re^{d}  
\label{eq1}
\end{equation}
where $d := d_e + d_r$ and $[a;b]_{1d}$ denote vector concatenation of vectors $a$ and $b$. The concatenated embedding is the input to all the subsequent layers.

\subsection{Common Feature Extraction Layer}
The common feature extraction is performed by a dense layer as well. This layer is commonly applied to all types of input embeddings. The width of a dense layer corresponds to the number of filters of the layer where each filter contains kernel size equal to that of input embeddings. The width is the key design parameter, in which we explain in more details in the experiment section.  

In the shared dense layer, we first apply an affine function $\Omega(\cdot)$ to input embeddings given by 
\begin{equation}
\Omega(x) = {W_h} x + b_h
\label{eq2}
\end{equation}
where $W_h \in \Re^{n d_h \times d}$ and $b_h \in \Re^{nd_h}$. The width of dense layer is given by $nd_h$, a multiple of $d_h$ where $n$ is the hyperparameter to be determined. The output of common feature extraction is obtained by applying the nonlinear activation $f(\cdot)$ to $\Omega([e_s;e_r]_{1d})$.

\subsection{Relation-aware Feature Extraction Layer}
In order to extract relation-specific features from concatenated embeddings (\ref{eq1}), we consider the relation-aware encoding function. The encoding function is denoted by $\Omega_r$ for relation $r$. Note that, if any triple contains the relation given by $r$, $\Omega_r(\cdot)$ is applied to input embeddings irrespective of subject or object entities. 
In ComDensE, we use a dense layer for the relation-aware feature extraction. The encoding function $\Omega_r$ is an affine function given by
\begin{equation}
\Omega_r(x) = {W_r}x + b_r
\label{eq3}
\end{equation}
where $W_r \in \Re^{d_z\times d } $, $b_r \in \Re^{d_z}$ and $d_z$ is the output length of $\Omega_r$. We first apply $\Omega_r$ to input embeddings (\ref{eq1}), and then apply the nonlinear activation function $f(\cdot)$.

\subsection{Projection to Embedding Space}
After obtaining the latent vectors from relation-aware and common feature extraction layers, the vectors are concatenated. The concatenated vector is projected to the embedding space by projection matrix $W \in \Re^{(d_z+nd) \times d_e}$, and then nonlinear activation $f(\cdot)$ is applied. Specifically, let us define $h_{sr} \in \Re^{d_e}$ as 
\begin{equation}
h_{sr} := f([f(\Omega_r([e_s;e_r]_{1d})); f(\Omega([e_s;e_r]_{1d}))]_{1d}^TW)
\label{eq4}
\end{equation}
The link prediction score $\psi(e_s, e_r, e_o)$ is defined as the inner product $h_{sr}^Te_o$. See Table \ref{table1} for the comparison of scoring function $\psi$. 

\subsection{Loss Function}
We compute the scores for all the triple, and calculate the loss using binary cross entropy function. We use 1:N training strategy introduced by~\cite{dettmers2018convolutional}. Let $N$ denote the number of all entities in KG. The loss $\mathcal{L}$ is given by
\begin{equation}
\mathcal{L} = -{1 \over N} \sum_{i=1}^N [(y_i) \log (p_i) + (1-y_i) \log (1-p_i)]
\label{eq5}
\end{equation}
where $p_i = \sigma(h_{sr}^Te_o^{(i)})$, $e_o^{(i)}$ is the $i$-th object entity, $y_i \in \left\{ 0,1 \right\}$ is the label, and $\sigma$ denote the sigmoid function.

\subsection{Properties of ComDensE model}

\noindent\textbf{Relation-specific inductive bias is important.} Typically in KGs, the number of relations is less than the number of entities, e.g., the ratio is 0.03\% for WN18RR, 1.6\% in FB15k-237, 0.03\% in YAGO3-10. This implies that triples in KGs having different relations tend to share multiple entities. Thus, relational inductive bias is needed so as to capture discriminative features of each relation in the presence of a large number of entities shared among different relations. ComDensE leverages both relation-aware and common feature extraction from embeddings to get the best of both approaches.

\noindent\textbf{Full-sized filters for concatenated embeddings are effective.} It was noted earlier that, it is important to mix as many components from different types of embeddings, i.e., entity and relation types when filtering the concatenated embeddings. Specifically in InteractE~\cite{vashishth2020interacte}, the authors propose the concept of  {\em heterogeneous interaction}. The heterogeneous interaction is defined as the degree to which a local (convolutional) filter can spatially cover different types, i.e., entity and relation types, of embedding components in the 2-D shaped concatenated embeddings. InteractE showed that the link prediction performance can be improved by increasing the heterogeneous interaction. Clearly, ComDensE maximizes such heterogeneous interaction by construction, because its dense layers use full-sized filters which entirely cover the 2-D area of the concatenated (and reshaped) embedding vectors of entities and relations.

\noindent\textbf{Computational overhead is manageable.} One of the merits of convolutional layers is parameter efficiency.  Indeed, the parameter count of dense layers is higher than the convolutional layers with smaller kernel sizes. We however argue that this incurs manageable computational overhead for KG embeddings. The size of KG embeddings are on the order of hundreds, which is much smaller than high-resolution image features. Furthermore, we later increase the width to capture abundant relational information for improving link prediction accuracy. In practice, prediction accuracy is often prioritized over parameter efficiency. Experiments also show that DensE incurs reasonable parameter counts compared to previous methods.

\begin{table}
\centering
\caption{Parameter sizes of InteractE and ComDensE.}
\label{table3}
\begin{tabular}{c|c|c}
\hline
\multicolumn{3}{c}{\textbf{The number of parameters}}\\
\hline
 \textbf{method} & \textbf{FB15k-237} & \textbf{WN18RR}\\
 \hline
 InteractE & 18M & 60M \\
  \hline
 ComDensE \textbf{256 $\times d$} & 66M & 33M \\
\hline
 ComDensE \textbf{128 $\times d$} & 35M & 22M \\
\hline
\end{tabular}
\end{table}

\begin{table}
\centering
\caption{All other hyperparameters are same except row dimension of relation-aware and common dense matrices. $d$ denote column dimension of dense matrices. While ComDensE uses 256 $\times d$ matrices to both benchmark datasets, ComDensE using 128 $\times d$ matrices achieves a similar performance gain on FB15k-237.
}
\label{table4}
\begin{tabular}{c|c|c}
\hline
\multicolumn{3}{c}{\textbf{Various sizes of matrix in ComDensE}}\\
\hline
 & \textbf{FB15k-237} & \textbf{WN18RR}\\
 \hline
 \multicolumn{3}{c}{\textbf{256$\times d$}}\\
 \textbf{MRR} & .356 & .473 \\
 \textbf{HIT@10} & .536 & .538 \\
 \textbf{HIT@1} & .265 & .440\\
  \hline
 \multicolumn{3}{c}{\textbf{128$\times d$}}\\
 \textbf{MRR} & .354 & .453 \\
 \textbf{HIT@10} & .534 & .512 \\
 \textbf{HIT@1} & .263 & .423 \\
\hline
\end{tabular}
\end{table} 

\begin{table*}[t]
\centering
\caption{Link prediction results of baseline models. \textbf{\underline{Bold with underline}} : the best performance, \textbf{bold} : the second best performance.}
\label{table5}
\begin{tabular}{cccc|ccc}
\hline
 & \multicolumn{3}{c}{\textbf{FB15k-237}} & \multicolumn{3}{c}{\textbf{WN18RR}}\\
 \hline
 \textbf{Methods} &
 \textbf{MRR} &
 \textbf{HIT@10} &
 \textbf{HIT@1} &
 \textbf{MRR} &
 \textbf{HIT@10} &
 \textbf{HIT@1} \\
 \hline
 TransE \cite{bordes2013translating} & .279 & .441  & - & .243 & .532 & -\\
 RotatE \cite{sun2019rotate} & .338 & .533 & .241 & \textbf{\underline{.476}} & \textbf{\underline{.571}} & .428\\
 \hline
 DistMult \cite{yang2014embedding} & .241 & .419 & .155 & .430 & .49 & .39\\
 ComplEx \cite{trouillon2016complex} & .247 & .428 & .158 & .44 & .51 & .41\\
 \hline
 RGCN \cite{schlichtkrull2018modeling} & .248 & .417 & .151 & - & - & -\\
 \hline
 ConvE \cite{dettmers2018convolutional} & .325 &  .501 & .237 & .43 & .52 & .40\\
 ConvTransE \cite{shang2019end} & .33 & .51 & .24 & .46 & .52 & .43\\
 SACN \cite{shang2019end} & .35 &  \textbf{\underline{.54}} & .26 & .47 & \textbf{.54} & .43\\
 InteractE \cite{vashishth2020interacte} & \textbf{.354} & .535 & \textbf{.263} & .463 & .528 & \textbf{.430}\\
 \hline
 ComDensE (proposed) & \textbf{\underline{.356}}  &  {\textbf{.536}} & \textbf{\underline{.265}} & \textbf{.473}  & .538 & \textbf{\underline{.440}}\\
 \hline
\end{tabular}
\end{table*}

\begin{table*}
\centering
\caption{Link prediction results by relation type on FB15k-237 dataset for ConvE, InteractE, and ComDensE. Relations are categorized into one-to-one (1:1), one-to-many (1:N), many-to-one (N:1) and many-to-many (N:N). We observe that ComDensE is effective in capturing most relations compared to baselines.}
\label{table6}
\begin{tabular}{cc|ccc|ccc|ccc}
\hline
  \multicolumn{2}{c}{} & \multicolumn{3}{c}{\textbf{ConvE}} & \multicolumn{3}{c}{\textbf{InteractE}} &  \multicolumn{3}{c}{\textbf{ComDensE}}\\
 \hline
 \multicolumn{2}{c}{\textbf{Relation Type}} &
 \textbf{MRR} &
 \textbf{HIT@10} &
 \textbf{HIT@1} &
 \textbf{MRR} &
 \textbf{HIT@10} &
 \textbf{HIT@1} &
 \textbf{MRR} &
 \textbf{HIT@10} &
 \textbf{HIT@1} \\
 \hline
 & \textbf{1:1} & .374 &.505 & .068 & .386 & .547 & .245 & \textbf{.422}  & \textbf{.557} & \textbf{.349}\\
 Pred & \textbf{1:N} & .095 & .17 & .030 & \textbf{.106} & \textbf{.192} & \textbf{.043} & .084 & .181 & \textbf{.043}\\
 Head & \textbf{N:1} & .444 & .644 & .315 & \textbf{.466} & .647 & .369 & \textbf{.466} & \textbf{.649} & \textbf{.372}\\
 & \textbf{N:N} & .261 & .459 & .140 & .276 & \textbf{.476} & .164 & \textbf{.279} & \textbf{.476} & \textbf{.187}\\
 \hline
 & \textbf{1:1} & .366 & .51 & .052 & .368 & .547 & .229  & \textbf{.422} & \textbf{.563} & \textbf{.349}\\
 Pred & \textbf{1:N} & .762 & .878 & .658 & .777 & .708 & \textbf{.881} & \textbf{.779} & \textbf{.884} & .717\\
 Tail & \textbf{N:1} & .069 &.15 & .015 & .074 &.141 & .034 & \textbf{.084} & \textbf{.169} & \textbf{.043}\\
 & \textbf{N:N} & .375 & .603 & .237 & .395 & .617 & .272 & \textbf{.396} & \textbf{.618} & \textbf{.285}\\
 \hline
\end{tabular}
\end{table*}

\section{Experiment}

\subsection{Datasets and Evaluation Settings}
ComDensE is evaluated on two KG benchmark datasets: FB15k-237 \cite{toutanova2015observed} and WN18RR \cite{dettmers2018convolutional}. These datasets are subsets of FB15k \cite{bordes2013translating} and WN18 \cite{bordes2013translating}, respectively. Specifically, inverse relations are removed from the original datasets so as to prevent direct inference via reversing triples. The statistics of benchmark datasets are summarized in Table \ref{table2}.

We evaluate the performance of link prediction in the filtered setting. Specifically, we compute scores from all other candidate triples in the test triples not existing in the training, validation, or test set, where candidates are generated by corrupting subjects for objects for predicting object. We use Mean Reciprocal Rank (MRR) and Hits at N (HIT@N) which are standard evaluation metrics for these datasets and models are evaluated in our experiments. We train and evaluate 5 times and then average performance results for robust evaluation. We compare ComDensE to baseline models: TransE \cite{bordes2013translating}, RotatE \cite{sun2019rotate}, DistMult \cite{yang2014embedding}, ComplEx \cite{trouillon2016complex}, RGCN \cite{schlichtkrull2018modeling}, ConvE \cite{dettmers2018convolutional}, ConvTransE \cite{shang2019end}, SACN \cite{shang2019end} and InteractE \cite{vashishth2020interacte}

\subsection{Hyperparameter Setting}
We use Adam~\cite{kingma2014adam} optimizer and search the hyperparameters on the validation set. The ranges of the hyperparameters for the grid design space: (embedding dimension) $\in$ \{150, 200, 256, 300\}, (batch size) $\in$ \{128, 256\}, (learning rate) $\in$ \{0.001, 0.0001\}, (input dropout) $\in$ \{0.4, 0.5\}, (hidden dropout) $\in$ \{0.4, 0.5\}, (width $n \times $ dense matrix $W^h \in \Re^{nd_h \times d}$, $n) \in$ \{2, 5, 50, 100, 200\}.\\
\indent Finally, we use same hyperparameters as embedding dimension 256, batch size 128, learning rate 0.0001, input dropout 0.4, row dimension of matrix $d_h$ 256 and hidden dropout 0.5 for both FB15k-237 and WN18RR. However, since the width is a crucial element of designing ComDensE and it depends on dataset sensitively, we only search a different width, where our method contains width 2 $\times$ square matrix $W^h \in \Re^{2d_h \times d}$ for FB15k-237 and width 100 $\times$ square matrix $W^h \in \Re^{100d_h \times d}$ for WN18RR, where $d_e$ denotes the dimension of entity embedding, $d_r$ denotes the dimension of relation embedding and $d := d_e + d_r$.

\section{Results}

\subsection{Prediction performance}
Table \ref{table5} demonstrates that ComDensE achieved the best link prediction performances in MRR, HIT@1 and competitive HIT@10 between entire KG embedding methods. Scores of baseline methods were taken from the respective works. In both FB15k-237 and WN18RR, ComDensE outperforms InteractE, which is the previous neural network based SOTA. For details; 0.6\% performance gain on MRR, 0.2\% HIT@10 and 1.1\% HIT@1 on FB15k-237. In addition, there also exist performance gain in WN18RR; 2.4\% performance gain in MRR, 2.5\% HIT@10, and 2.6\% in HIT@1. This simple method outperformed shared-layer models like ConvE, SACN, InteractE and a relation-specific model like RGCN. Our result demonstrates that the architecture combining two different encoding functions improves performance of the KG embedding.

\subsection{Exploring parameter efficiency}
One aspect of ComDensE is that, since it mostly uses dense layers, it may be less parameter-efficient than CNN. While the parameter counts of ComDensE for FB15k-237 were higher than InteractE, those were less than InteractE for WN18RR datasets. Table \ref{table3} and Table \ref{table4} demonstrates experimental results of ComDensE using 256 $\times d$ and 128 $\times d$  matrices.

\subsection{Performance with Different Relation Types}
We analyzed the performance of ComDensE on different relation types of FB15k-237, since FB15k-237 contains more diverse relations than WN18RR. Four relation types based on the number of tails connected with head and the number of heads connected with tail were defined by~\cite{wang2014knowledge}: one-to-one (1:1), one-to-many (1:N), many-to-one (N:1), and many-to-many (N:N). Using this dataset, we compared three models: ConvE, InteractE, and ComDensE at four types of relations. The results are presented in Table \ref{table6}. We find that ComDensE is effective in complex relation types, i.e.,  1:N, N:N, N:1, as well as simple relations, i.e., 1:1. Notably, the performance gain is higher in 1:1, demonstrating that ComDensE particularly effective in capturing simple relationships.



\begin{table*}
\centering
\caption{Ablation study: effect of the width of shared dense layer-only models. Width $n\times$ represent the width of $W_h$ is $nd$, i.e.,  $W_h \in \Re^{nd\times d}$}
\label{table7}
\begin{tabular}{cccc|ccc}
\hline
 & \multicolumn{3}{c}{\textbf{FB15k-237}} & \multicolumn{3}{c}{\textbf{WN18RR}}\\
 \hline
 \textbf{Methods} &
 \textbf{MRR} &
 \textbf{HIT@10} &
 \textbf{HIT@1} &
 \textbf{MRR} &
 \textbf{HIT@10} &
 \textbf{HIT@1} \\
 \hline
 Width 1 $\times$ (baseline) & .336 & .515  & .246 & .437 & .503 & .402\\
 \hline
 Width 50 $\times$ & .348 \textbf{(+.012)} & .532 \textbf{(+.017)} & .258 \textbf{(+.012)} & .471 \textbf{(+.034)} & .537 \textbf{(+.034)}& .439 \textbf{(+.037)}\\
 Width 100 $\times$ & .349 \textbf{(+.013)} & .531 \textbf{(+.016)} & .260 \textbf{(+.014)} & .475 \textbf{(+.038)} & .543 \textbf{(+.04)}  & .441 \textbf{(+.039)}\\
 Width 200 $\times$ & .347 \textbf{(+.011)} & .529 \textbf{(+.014)} & .258 \textbf{(+.012)} & \textbf{.476} \textbf{(+.039)}& \textbf{.545} \textbf{(+.042)}& \textbf{.443} \textbf{(+.041)}\\
 Width 300 $\times$ & .347 \textbf{(+.011)} & .529 \textbf{(+.014)} & .257 \textbf{(+.011)} & .462 \textbf{(+.025)} & .525 \textbf{(+.022)} & .431 \textbf{(+.029)}\\
 \hline
  ComDensE & \textbf{.356 (+.02)}  &  \textbf{.536 (+.021)} & \textbf{.265 (+.019)} & .473 \textbf{(+.036)}  & .538 \textbf{(+.035)} & .440 \textbf{(+.038)}\\
  \hline
\end{tabular}
\end{table*}

\begin{table*}
\centering
\caption{Ablation study: effect of the depth of shared dense layer-only models. Depth-$m$ denote the depth $m$ of common dense layer.}
\label{table8}
\begin{tabular}{cccc|ccc}
\hline
 & \multicolumn{3}{c}{\textbf{FB15k-237}} & \multicolumn{3}{c}{\textbf{WN18RR}}\\
 \hline
 \textbf{Methods} &
 \textbf{MRR} &
 \textbf{HIT@10} &
 \textbf{HIT@1} &
 \textbf{MRR} &
 \textbf{HIT@10} &
 \textbf{HIT@1} \\
 \hline
 Depth-1 (baseline) & .336 & \textbf{.515}  & .246 & \textbf{.437} & \textbf{.503} & \textbf{.402}\\
 \hline
 Depth-2 & \textbf{.338} \textbf{(+.002)} & \textbf{.515} & \textbf{.250} \textbf{(+.004)} & .426 \textbf{(-.011)} & \textbf{.503} & .380 \textbf{(-.022)}\\
 Depth-3 & .328 \textbf{(-.008)} & .499 \textbf{(-.016)} & .242 \textbf{(-.004)} & .446 \textbf{(+.009)} & .510 \textbf{(+.007)} & .369 \textbf{(-.033)}\\
 Depth-4 & .313 \textbf{(-.023)} & .478 \textbf{(-.037)} & .230 \textbf{(-.016)} & .382 \textbf{(-.055)} & .491 \textbf{(-.012)} & .320 \textbf{(-.082)}\\
 Depth-5 & .301 \textbf{(-.035)} & .460 \textbf{(-.055)} & .220 \textbf{(-.026)} & .339 \textbf{(-.098)} & .465 \textbf{(-.038)} & .273 \textbf{(-.129)}\\
 \hline
  ComDensE & \textbf{.356 (+.02)}  &  \textbf{.536 (+.021)} & \textbf{.265 (+.019)} & \textbf{.473 (+.036)}  & \textbf{.538 (+.035)} & \textbf{.440 (+.038)}\\
 \hline
\end{tabular}
\end{table*}

\section{Ablation Study}
This section conducted ablation study associated with ComDensE. Since ComDensE combines both relation-aware and shared dense layer, we studied models in which some of the layer is removed or changed. Specifically, we considered the following configurations:
\begin{itemize}
		\item Shared dense layer only with varying widths
		\item Shared dense layer only with varying depths
		\item Different relation-aware encoding functions
	\end{itemize}
	
\subsection{Shared dense layer only: Effects of Widths}
We evaluated ComDensE only containing the shared dense layer, i.e., removing the relation-aware dense layer. We adjusted the shape of the shared dense layer in order to see if the model can outperform original ComDensE. We first explored the effects of the width of shared dense layer. Note that, increasing the width of dense layer is equivalent to increasing the number of kernels in CNN. As in Table \ref{table7}, the optimal widths are roughly $100d$ and $200d$ respectively for FB15k-237 and WN18RR models. Thus, it appeared that increasing the number of filters indefinitely does not help. In most cases, ComDensE outperformed other configurations, whereas in WN18RR, ComDensE was slightly worse, while the performance gap was quite small.

\subsection{Shared dense layer only: Effects of Depths}
Next, we also considered shared dense layer only and evaluated the effect of increasing network depth. We increased the number of layers in the shared network where the depth is in $\{2, 3, 4, 5\}$. In general, increasing the depth enhances expressive power of the model, while it did not apply to our experiments. Table \ref{table8} illustrates that link prediction performance aggravates with increasing depth. Thus, we concluded that the width of ComDensE is the only important design parameter than the depth in order to obtain a better representation of KGs.

\subsection{Different relation-aware encoding functions}
This experiment considers the effect of different relation-aware encoding functions $\Omega_r$. We tested a simple ``translation-only'' function such that
\begin{equation}
\Omega_r(x) = x + v_r
\label{eq9}
\end{equation}
where $v_r \in \Re^{d}$ is the learnable relation-specific vectors. The results and comparison with ConvE were illustrated in Table \ref{table10}. Note ``translation-only'' $\Omega_r$ does not provide any heterogeneous interaction between embedding components. However, the link prediction performance was better than ConvE on both FB15k-237 and WN18RR. This demonstrated that the inductive bias from relation-awareness enhances the link prediction performance. However, the results in Table \ref{table10} are worse than ComDensE, which shows that promoting the heterogeneous interactions~\cite{vashishth2020interacte} among the embedding components is crucial.

\begin{table}[h]
\centering
\caption{Prediction performance when $\Omega_r$ is ``translation only'' function, i.e., vector addition. (+,-) denote the performance difference from ConvE~\cite{dettmers2018convolutional}}
\label{table10}
\begin{tabular}{c|c|c}
\hline
\multicolumn{3}{c}{\textbf{vector add operation}}\\
\hline
 & \textbf{FB15k-237} & \textbf{WN18RR}\\
 \hline
 \textbf{MRR} & .344 (+.019) & .460 (+.030) \\
 \textbf{HIT@10} & .527 (+.026) & .522 (+.002) \\
 \textbf{HIT@1} & .253 (+.016) & .428 (+.028)\\
\hline
\end{tabular}
\end{table}

\section{Conclusion}
We propose ComDensE, which combines two parallel models consisting mainly of \emph{dense} layers: one is \emph{relation-aware encoder} considering inductive bias based on relations and the other is \emph{common feature extractor} for all the triples irrespective of relations. In our extensive experiments, ComDensE achieves state-of-the-art performance for link prediction on both FB15k-237 \textbf{(MRR, HIT@1)} and WN18RR \textbf{(HIT@1)}. In addition, our ablation study demonstrates that the relation-aware encoder is essential for completing high-degree KGs like FB15k-237 and fully-connected common feature extraction improves the prediction for the low-degree KGs like WN18RR. Notably, increasing width to a certain extent improves link prediction performance, however, increasing depth is deemed harmful. In the future, we plan to introduce architectural novelty to the model inspired by other disciplines such as computer vision or natural language processing.  

\bibliographystyle{IEEEtran}
\bibliography{paper}

\end{document}